\documentclass[10pt,twocolumn,letterpaper]{article}
\pdfoutput=1

\usepackage{cvpr}
\usepackage{times}
\usepackage{epsfig}
\usepackage{graphicx}
\usepackage{amsmath}
\usepackage{amssymb}

\usepackage[T1]{fontenc}

\usepackage{multirow}
\usepackage{tabularx}
\usepackage{booktabs}
\usepackage[table]{xcolor}
\usepackage{url}

\usepackage{algorithm}
\usepackage{algorithmicx}
\usepackage[noend]{algpseudocode}
\usepackage[numbers,sort]{natbib}

\usepackage{widetext}

\graphicspath{{pix/}{plots/}}

\newcommand{\gc}{\cellcolor{gray!25}}

\usepackage[pagebackref=true,breaklinks=true,letterpaper=true,colorlinks,bookmarks=false]{hyperref}

\usepackage{subfigure}

\cvprfinalcopy 

\def\cvprPaperID{573} 

\ifcvprfinal\fi

\usepackage[first=0,last=9]{lcg}

\usepackage{xspace}
\usepackage{bbold}
\usepackage{xcolor}

\newcommand{\set}[1]{\ensuremath{\mathcal{#1}}}

\newcommand{\bnStd}{\textsc{Std-BN}\xspace}
\newcommand{\bnInplace}{\textsc{InPlace-ABN}\xspace}
\newcommand{\bnInplaceSync}{\textsc{InPlace-ABN$^{\mathsf{sync}}$}\xspace}
\newcommand{\fixedcrop}{\textsc{fixed crop}\xspace}
\newcommand{\fixedbatch}{\textsc{fixed batch}\xspace}

\usepackage{enumitem}
\setitemize{leftmargin=*,itemsep=0pt}

\newcommand{\myparagraph}[1]{\vspace{5pt}\noindent\textbf{#1}}

\begin{document}

\title{In-Place Activated BatchNorm for Memory-Optimized Training of DNNs}

\author{Samuel Rota Bul\`o, Lorenzo Porzi, Peter Kontschieder\\
Mapillary Research\\
{\tt\small research@mapillary.com}
}

\maketitle

\begin{abstract}
In this work we present In-Place Activated Batch Normalization (\bnInplace) -- a novel approach to drastically reduce the training memory footprint of modern deep neural networks in a computationally efficient way. Our solution substitutes the conventionally used succession of BatchNorm + Activation layers with a single plugin layer, hence avoiding invasive framework surgery while providing straightforward applicability for existing deep learning frameworks. We obtain memory savings of up to 50\% by dropping intermediate results and by recovering required information during the backward pass through the inversion of stored forward results, with only minor increase ($0.8$-$2\%$) in computation time. Also, we demonstrate how frequently used checkpointing approaches can be made computationally as efficient as \bnInplace. In our experiments on image classification, we demonstrate on-par results on ImageNet-1k with state-of-the-art approaches. On the memory-demanding task of semantic segmentation, we report results for COCO-Stuff, Cityscapes and Mapillary Vistas, obtaining new state-of-the-art results on the latter without additional training data but in a single-scale and -model scenario. Code can be found at \url{https://github.com/mapillary/inplace_abn} .
\end{abstract}


\section{Introduction}
High-performance computer vision recognition models typically take advantage of deep network backbones, generating rich feature representations for target applications to operate on. For example, top-ranked architectures used in the 2017 LSUN or MS COCO segmentation/detection challenges are predominantly based on ResNet/ResNeXt~\cite{He2015b, Xie2016} models comprising >100 layers. 

Obviously, depth/width of networks strongly correlate with GPU memory requirements and at given hardware memory limitations, trade-offs have to be made to balance feature extractor performance vs.~application-specific parameters like network output resolution or training data size. A particularly memory-demanding task is semantic segmentation, where one has to compromise significantly on the number of training crops per minibatch and their spatial resolution. In fact, many recent works based on modern backbone networks have to set the training batch size to no more than \textit{a single} crop per GPU~\cite{Wu2016,Chen2016}, which is partially also due to suboptimal memory management in some deep learning frameworks. In this work, we focus on increasing the memory efficiency of the training process of modern network architectures in order to further leverage performance of deep neural networks in tasks like image classification and semantic segmentation.
\begin{figure}[t]
 \centering
 \includegraphics{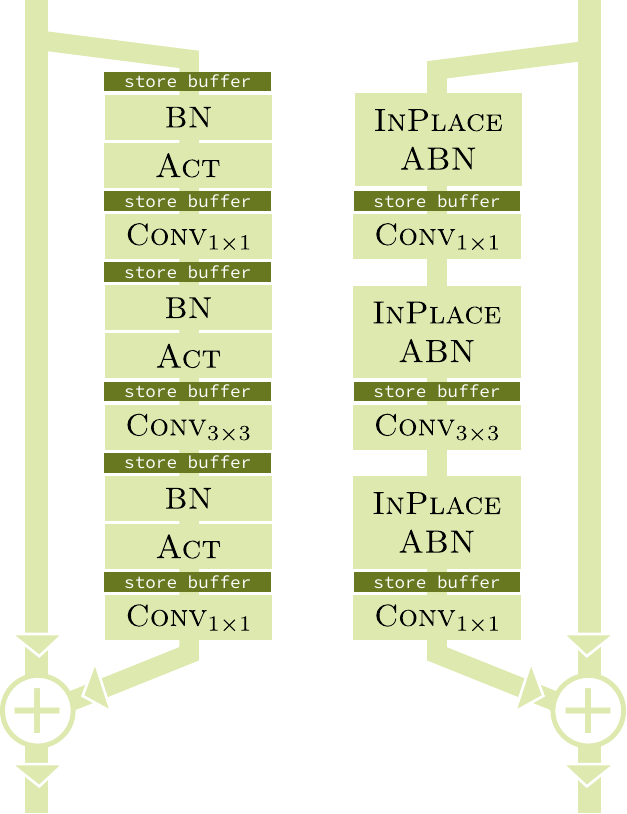}
 \caption{Example of residual block with identity mapping~\cite{He+16}. Left: Implementation with standard \textsc{BN} and in-place activation layers, which requires storing $6$ buffers for the backward pass. Right: Implementation with our proposed \bnInplace layer, which requires storing only $3$ buffers. Our solution avoids storing the buffers that are typically kept for the backward pass through \textsc{BN} and exhibits a lower computational overhead compared to state-of-the-art memory-reduction methods.}
 \label{fig:teaser}
 \vspace{-12pt}
\end{figure}
 
We introduce a novel and unified layer that replaces the commonly used succession of batch normalization (\textsc{BN}) and nonlinear activation layers (\textsc{Act}), which are integral with modern deep learning architectures like ResNet~\cite{He2015b}, ResNeXt~\cite{Xie2016}, Inception-ResNet~\cite{Szegedy2016}, WideResNet~\cite{Zagoruyko2016WRN}, Squeeze-and-Excitation Networks~\cite{Hu2017}, DenseNet~\cite{Huang2017}, \etc. Our solution is coined \bnInplace and proposes to merge batch normalization and activation layers in order to enable in-place computation, using only a single memory buffer for storing the results (see illustration in Figure~\ref{fig:teaser}). During the backward pass, we can efficiently recover all required quantities from this buffer by inverting the forward pass computations.
Our approach yields a theoretical memory reduction of up to $50\%$, and our experiments on semantic segmentation show additional data throughput of up to $+75\%$ during training, when compared to prevailing sequential execution of \textsc{BN+Act}. Our memory gains are obtained without introducing noticeable computational overhead, \ie~side-by-side runtime comparisons show only between +$0.8$-$2\%$ increase in computation time. 

As additional contribution, we review the \textit{checkpointing} memory management strategy~\cite{Chen+16} and propose a computationally optimized application of this idea in the context of \textsc{BN} layers. This optimization allows us to drop re-computation of certain quantities needed during the backward pass, eventually leading to reduced computation times as per our \bnInplace. However, independent of the proposed optimized application of~\cite{Chen+16}, conventional checkpointing in general suffers from higher implementation complexity (with the necessity to invasively manipulate the computation graph), while our main \bnInplace contribution can be easily implemented as self-contained, standard \textit{plug-in layer} and therefore simply integrated in any modern deep learning framework.

Our experimental evaluations demonstrate on-par performance with state-of-the-art models trained for image classification on ImageNet~\cite{Rus+15} (in directly comparable memory settings), and significantly improved results for the memory-critical application of semantic segmentation. 

To summarize, we provide the following contributions:
\begin{itemize}
 \item Introduction of a novel, self-contained \bnInplace layer that enables joint, in-place computation of \textsc{BN+Act}, approximately halvening the memory requirements during training of modern deep learning models.
 \item A computationally more efficient application of the \textit{checkpointing} memory management strategy in the context of \textsc{BN} layers, inspired by optimizations used for \bnInplace.
 \item Experimental evaluations for i) image classification on ImageNet-1k showing approximately on-par performance with state-of-the-art models and ii) semantic segmentation on COCO-Stuff, Cityscapes and Mapillary Vistas, considerably benefiting from the additional available memory and generating new high-scores on the challenging Vistas dataset.
\end{itemize}

\section{Related Work}
The topic of optimizing memory management in deep learning frameworks is typically addressed at different levels. Efficient deep learning frameworks like TensorFlow, MxNet or PyTorch follow distinct memory allocation strategies. Among them is \textit{checkpointing}~\cite{Martens2012,Chen+16}, which provides additional memory at the cost of runtime via storing activation buffers as so-called \textit{checkpoints}, from where required quantities can be re-computed during the backward pass. The paper in~\cite{Chen+16} describes how to recursively apply such a variant on sub-graphs between checkpoints. In~\cite{Gruslys2016} this is further optimized with dynamic programming, where a storage policy is determined that minimizes the computational costs for re-computation at a fixed memory budget. 

Virtually all deep learning frameworks based on NVIDIA hardware exploit low-level functionality libraries CUDA and cuDNN\footnote{\url{https://developer.nvidia.com}}, providing GPU-accelerated and performance-optimized primitives and basic functionalities. Another line of research has focused on training CNNs with reduced precision and therefore smaller memory-footprint datatypes. Such works include (partially) binarized weights/activations/gradients~\cite{Courbariaux2015,Hubara2016a,Hubara2016b}, which however typically lead to degraded overall performance. With mixed precision training~\cite{Micikevicius2017}, this issue seems to be overcome and we plan to exploit this as complementary technique in future work, freeing up even more memory for training deep networks without sacrificing runtime.

In \cite{Gomez2017} the authors modify ResNet in a way to contain reversible residual blocks, \ie residual blocks whose activations can be reconstructed backwards. Backpropagation through reversible blocks can be performed without having stored intermediate activations during the forward pass, which allows to save memory. However, the cost to pay is twofold. First, one has to recompute each residual function during the backward pass, thus having the same overhead as checkpointing~\cite{Martens2012}. Second, the network design is limited to using blocks with certain restrictions, \ie reversible blocks cannot be generated for bottlenecks where information is supposed to be discarded. 

Finally, we stress that only training time memory-efficiency is targeted here while test-time optimization as done \eg in NVIDIAs TensorRT~\footnote{\url{https://developer.nvidia.com/tensorrt}} is beyond our scope.


\section{In-Place Activated Batch Normalization}
Here, we describe our contribution to avoid the storage of a buffer that is typically needed for the gradient computation during the backward pass through the batch normalization layer. As opposed to existing approaches we also show that our solution minimizes the computational overhead we have to trade for saving additional memory. 

\subsection{Batch Normalization Review}
Batch Normalization has been introduced in~\cite{IofSze15} as an effective tool to reduce internal covariate shift in deep networks and accelerate the training process.
Ever since, \textsc{BN} plays a key role in most modern deep learning architectures. 

The key idea consists in having a normalization layer that applies an axis-aligned whitening of the input distribution, followed by a scale-and-shift operation aiming at preserving the network's representation capacity.
The whitening operation exploits statistics computed on a minibatch level only. The by-product of this approximation is an additional regularizing effect for the training process.

In details, we can fix a particular unit $x$ in the network and let $x_\set B=\{x_1,\dots,x_m\}$ be the set of values $x$ takes from a minibatch $\set B$ of $m$ training examples. The batch normalization operation applied to $x_i$ first performs a whitening of the activation using statistics computed from the minibatch:
\begin{equation}
	\hat x_i=\textsc{BN}(x_i)=\frac{x_i-\mu_{\set B}}{\sqrt{\sigma^2_{\set B}+\epsilon}}\,.
	\label{eq:xhat}
\end{equation}
Here $\epsilon>0$ is a small constant that is introduced to prevent numerical issues, 
and $\mu_\set B$ and $\sigma^2_\set B$ are the empirical mean and variance of the activation unit $x$, respectively,  computed with respect to the minibatch $\set B$, \ie
\[
	\quad \mu_{\set B}=\frac{1}{m}\sum_{j=1}^m x_j\,,\qquad\sigma^2_{\set B}=\frac{1}{m}\sum_{j=1}^m(x_j-\mu_\set B)^2\,.
\]
The whitened activations $\hat x_i$ are then scaled and shifted by learnable parameters $\gamma$ and $\beta$, obtaining
\[
	y_i=\textsc{BN}_{\gamma,\beta}(x_i)=\gamma \hat x_i+\beta\,.
\]
The \textsc{BN} transformation described above can in principle be applied to any activation in the network and is typically adopted with channel-specific $(\gamma,\beta)$-parameters. Using \textsc{BN} 
renders training resilient to the scale of parameters, thus enabling the use of higher learning rates.

At test time, the \textsc{BN} statistics are fixed to $\mu_\set T$ and $\sigma_\set T$, estimated from the entire training set $\set T$. These statistics are typically updated at training time with a running mean over the respective minibatch statistics, but could also be recomputed before starting the testing phase. Also, the computation of networks trained with batch normalization can be sped up by absorbing \textsc{BN} parameters into the preceding \textsc{Conv} layer, by performing a simple update of the convolution weights and biases. This is possible because at test-time \textsc{BN} becomes a linear operation.

\subsection{Memory Optimization Strategies}\label{ss:overview}
Here we sketch our proposed memory optimization strategies after introducing both, the standard (memory-inefficient) use of batch normalization and the state-of-the-art coined \emph{checkpointing}~\cite{Martens2012,Chen+16}.
\newpage
\begin{figure}[h]
	\centering
	\subfigure[Standard building block (memory-inefficient)]{\label{sfig:standard}\includegraphics[width=.9\columnwidth]{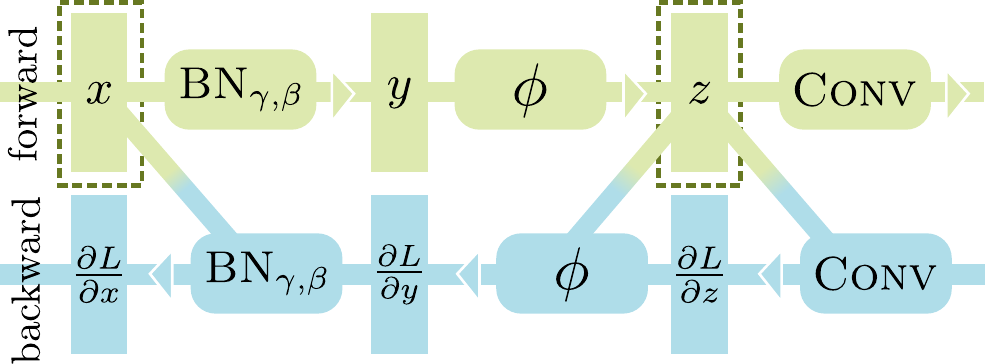}} %
	\subfigure[Checkpointing~\cite{Martens2012,Chen+16}]{\label{sfig:mxnet}\includegraphics[width=.9\columnwidth]{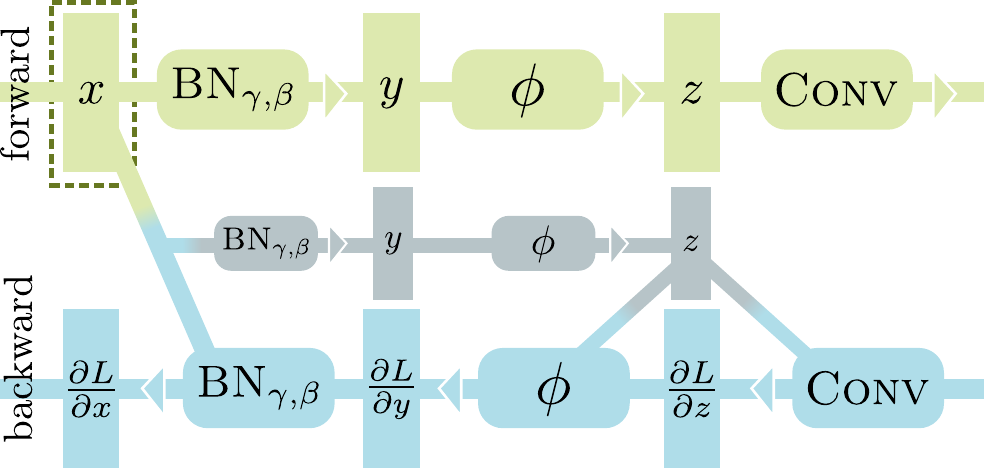}} %
	\subfigure[Checkpointing (proposed version)]{\label{sfig:mxnet_optim}\includegraphics[width=.9\columnwidth]{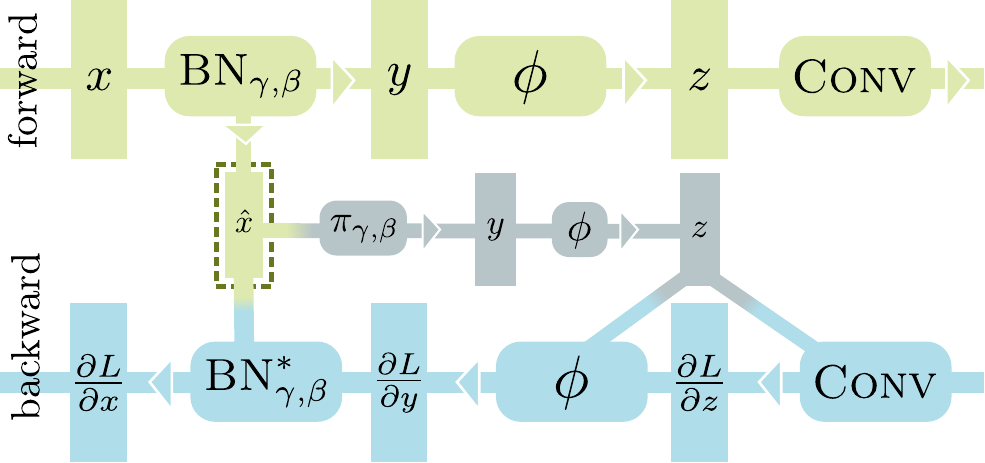}} %
	\subfigure[In-Place Activated Batch Normalization I (proposed method)]{\label{sfig:our}\includegraphics[width=.9\columnwidth]{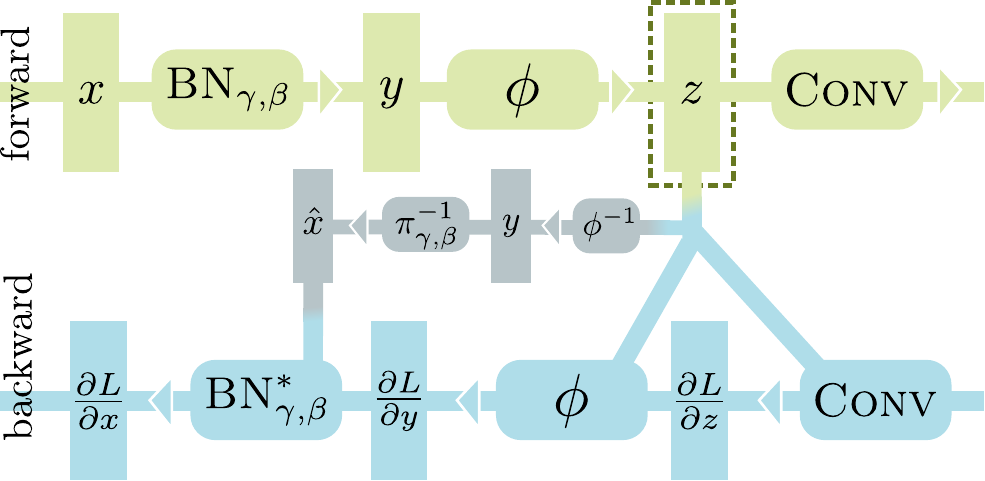}} %
	\subfigure[In-Place Activated Batch Normalization II (proposed method)]{\label{sfig:ourII}\includegraphics[width=.9\columnwidth]{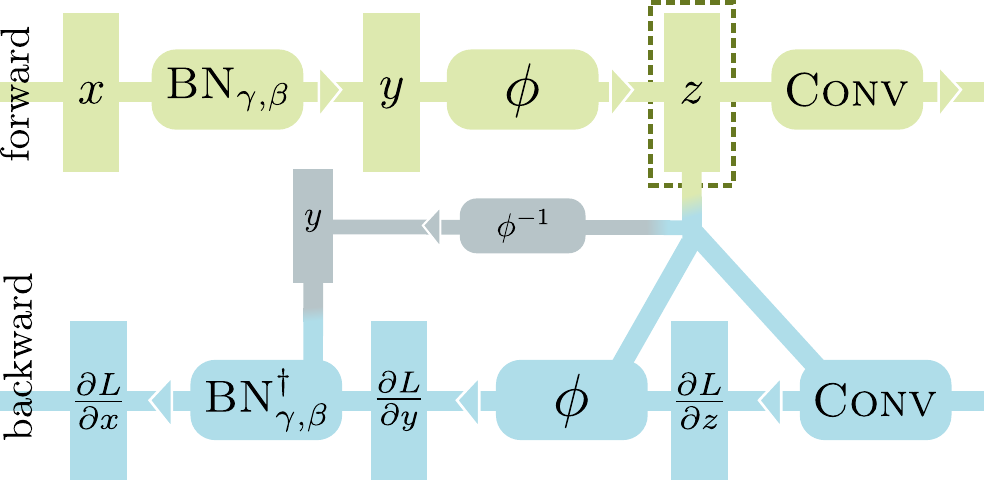}} %
	\caption{Comparison of standard \textsc{BN}, state-of-the-art checkpointing from~\cite{Martens2012,Chen+16} and our proposed methods. See \S~\ref{ss:overview} for a detailed description.}
	\label{fig:overview}
	\vspace{-15pt}
\end{figure}

In Figure~\ref{fig:overview}, we provide diagrams showing the forward and backward passes of a typical building block \textsc{BN+Act+Conv}\footnote{Having the convolution at the end of the block is not strictly necessary, but supports comprehension.} that we find in modern deep architectures. 
The activation function (\eg \textsc{ReLU}) is denoted by $\phi$.
Computations occurring during the forward pass are shown in green and involve the entire minibatch $\set B$ (we omit the subscript $\set B$). Computations happening during the backward pass are shown in cyan and gray. The gray part aims at better highlighting the additional computation that has been introduced to compensate for the memory savings.
Rectangles are in general volatile buffers holding intermediate results, except for rectangles surrounded by a dashed frame, which represent buffers that need to be stored for the backward pass and thus significantly impact the training memory footprint. \Eg, in Figure~\ref{sfig:standard} only $x$ and $z$ will be stored for the backward pass, while in Figure~\ref{sfig:mxnet} only $x$ is stored. 
For the sake of presentation clarity, we have omitted two additional buffers holding $\mu_\set B$ and $\sigma_\set B$ for the \text{BN} backward phase. Nevertheless, these buffers represent in general a small fraction of the total allocated memory. Moreover, we have also omitted the gradients with respect to the model parameters (\ie $\gamma$, $\beta$ and \textsc{Conv} weights).

\myparagraph{Standard.}
In Figure~\ref{sfig:standard} we present the standard implementation of the reference building block, as used in all deep learning frameworks. During the forward pass both, the input $x$ to \textsc{BN} and the output of the activation function $\phi$ need to be stored for the backward pass. Variable $x$ is used during the backward pass through $\textsc{BN}_{\gamma,\beta}$ to compute both the gradient \wrt its input and $\gamma$, \ie $\frac{\partial L}{\partial x}$ and $\frac{\partial L}{\partial\gamma}$ where $L$ denotes the loss, while $z$ is required for the backward pass through the activation $\phi$ as well as potential subsequent operations like \eg the convolution shown in the figure.


\myparagraph{Checkpointing~\cite{Martens2012,Chen+16}.}
This technique allows to trade computation for memory when training neural networks, applicable in a very broad setting. In Figure~\ref{sfig:mxnet}, we limit its application to the building block under consideration like in~\cite{Pleiss+17}.
In contrast to the standard implementation, which occupies two buffers for the backward pass of the shown building block,
checkpointing requires only a single buffer. 
The trick consists in storing only $x$ and recomputing $z$ during the backward pass by reiterating the forward operations starting from $x$ (see gray-colored operations). Clearly, the computational overhead to be paid comprises both, recomputation of the \textsc{BN} and activation layers. 
It is worth observing that recomputing $\textsc{BN}_{\gamma,\beta}$ (gray) during the backward phase can reuse values for $\mu_{\set B}$ and $\sigma_{\set B}$ available from the forward pass and fuse together the normalization and subsequent affine transformation into a single scale-and-shift operation. Accordingly, the cost of the second forward pass over $\textsc{BN}_{\gamma,\beta}$ becomes less expensive~(see also~\cite{Pleiss+17}).

\vspace{5pt}
\emph{The three approaches that follow are all contributions of this work. The first represents a variation of checkpointing, which allows us to save additional computations in the context of \textsc{BN}. The second and third are our main contributions, providing strategies that yield the same memory savings and even lower computational costs compared to the proposed, optimized checkpointing, but are both self-contained and thus much easier to integrate in existing deep learning frameworks.}

\myparagraph{Checkpointing (proposed version).}
Direct application of the checkpointing technique in the sketched building block, which is adopted also in~\cite{Pleiss+17}, is not computationally optimal since additional operations could be saved by storing $\hat x$, \ie the normalized value of $x$ as per Eq.~\eqref{eq:xhat}, instead of $x$. Indeed, as we will see in the next subsection, the backward pass through \textsc{BN} requires recomputing $\hat x$ if not already stored. For this reason, we propose in Figure~\ref{sfig:mxnet_optim} an alternative implementation that is computationally \emph{more efficient} by retaining $\hat x$ from the forward pass through the \textsc{BN} layer. From $\hat x$ we can recover $z$ during the backward pass by applying the scale-and-shift operation $\pi_{\gamma,\beta}(\hat x)=\gamma\hat x+\beta$, followed by the activation function $\phi$ (see gray-colored operations). In this way, the computation of $z$ becomes slightly more efficient than the one shown in Figure~\ref{sfig:mxnet}, for we save the fusion operation. Finally, an additional saving of the normalization step derives from using the stored $\hat x$ in the backward implementation of \textsc{BN} rather than recomputing it from $x$. To distinguish the efficient backward implementation of \textsc{BN} from the standard one we write $\textsc{BN}_{\gamma,\beta}^*$ in place of $\textsc{BN}_{\gamma,\beta}$ (cyan-colored, see additionally \S~\ref{ss:tenchnical}).

\myparagraph{In-Place Activated Batch Normalization I.}
A limitation of the memory-reduction strategy described above is that the last layer, namely $\textsc{Conv}$ in the example, depends on non-local quantities like $x$ (or $\hat x$) for the computation of the gradient. This makes the implementation of the approach within standard frameworks somewhat cumbersome, because the backward pass of any layer that follows $\phi$, which relies on the existence of $z$, has to somehow trigger its recomputation.
To render the implementation of the proposed memory savings \emph{easier} and \emph{self-contained}, we suggest an alternative strategy shown in Figure~\ref{sfig:our}, which relies on having only $z$ as the saved buffer during the forward pass, thus operating an in-place computation through the \textsc{BN} layer (therefrom the paper's title).
By doing so, any layer that follows the activation $\phi$ would have the information for the gradient computation locally available.
Having stored $z$, we need to recompute $\hat x$ \emph{backwards}, for it will be needed in the backward pass through the \textsc{BN} layer.\footnote{This solution can technically still be considered as a form of checkpointing, but instead of recovering information forwards as in~\cite{Martens2012,Chen+16}, we recover it backwards, thus bearing a similarity to reversible nets~\cite{Gomez2017}.}
However, this operation is only possible if the activation function is invertible. Even though this requirement does not hold for \textsc{ReLU}, \ie~one of the most dominantly used activation functions, we show in \S~\ref{ssec:ExpClass} that an invertible function like \textsc{Leaky ReLU}~\cite{MaaHanNg13} with a small slope works well as a surrogate of \textsc{ReLU} without compromising on the model quality.
We also need to invert the scale-and-shift operation $\pi_{\gamma,\beta}$, which is in general possible if $\gamma\neq 0$. 

\myparagraph{In-Place Activated Batch Normalization II.}
The complexity of the computation of $\hat x=\pi_{\gamma,\beta}^{-1}(y)=\frac{y-\beta}{\gamma}$ used in the backward pass of \bnInplace I can be further reduced by rewriting the gradients $\frac{\partial L}{\partial \gamma}$ and $\frac{\partial L}{\partial x}$ directly as functions of $y$ instead of $\hat x$. The explicit inversion of $\pi_{\gamma,\beta}$ to recover $\hat x$ applies $m$ scale-and-shift operations (per feature channel). If the partial derivatives are however based on $y$ directly, the resulting modified gradients (derivations given in the Appendix) show that the same computation can be absorbed into the gradient $\frac{\partial L}{\partial x_i}$ at $\mathcal O(1)$ cost (per feature channel). In Figure~\ref{sfig:ourII} we show the diagram of this optimization, where we denote as $\textsc {BN}^\dagger_{\gamma,\beta}$ the implementation of the backward pass as a function of $y$. 

\subsection{Technical Details}\label{ss:tenchnical}
The key components of our method are the computation of the inverse of both the activation function (\bnInplace I \& II) and $\pi_{\gamma,\beta}$ (\bnInplace I), and the implementation of a backward pass through the batch normalization layer that depends on $y$, \ie the output of the forward pass through the same layer.

\begin{figure}[htb]
	\centering
	\includegraphics[width=.49\columnwidth]{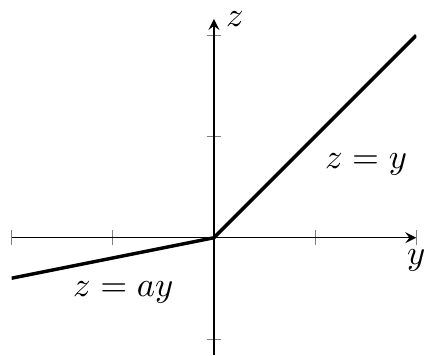}
	\includegraphics[width=.49\columnwidth]{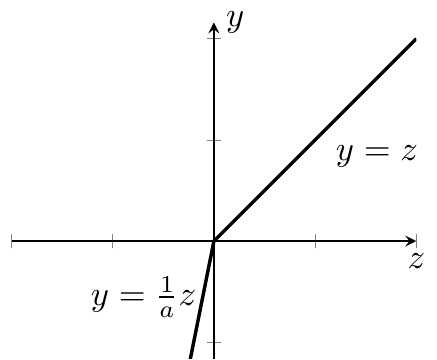}
	\caption{\textsc{Leaky ReLU} with slope $a$ (left) and its inverse (right).}
	\label{fig:lrelu}
	\vspace{-5pt}
\end{figure}

\myparagraph{Invertible activation function.} Many activation functions are actually invertible and can be computed in-place (\eg sigmoid, hyperbolic tangent, \textsc{Leaky ReLU}, and others), but the probably most commonly used one, namely \textsc{ReLU}, is not invertible. However, we can replace it with \textsc{Leaky ReLU} (see, Figure~\ref{fig:lrelu}) with slope $0.01$ without impacting the quality of the trained models~\cite{Xu+15}. This will be the activation function that we use in our experimental evaluation due to its affinity to standard \textsc{ReLU}, even though other activation functions could be used.
The corresponding forward pass through the activation function with slope $a$ for negative inputs and its inverse used in our backward pass are given as follows:
\[
	f(y)=
\begin{cases}
	y&\text{if }y\geq 0\\
	ay&\text{if }y<0
\end{cases},\quad
f^{-1}(z)=
\begin{cases}
	z&\text{if }z\geq 0\\
	\frac{z}{a}&\text{if }z<0\
\end{cases}\,.
\]
\textsc{Leaky ReLU} and its inverse share the same computational cost, \ie an elementwise sign check and scaling operation. Hence, the overhead deriving from the recomputation of $\phi$ in the backward pass of the previously shown, checkpointing-based approaches and its inverse $\phi^{-1}$ employed in the backward pass of our method are equivalent. To give further evidence of the interchangeability of \textsc{ReLU} and \textsc{Leaky ReLU} with slope $a=0.01$, we have successfully retrained well-known models like ResNeXt and WideResNet on ImageNet using \textsc{Leaky ReLU} (see \S~\ref{ssec:ExpClass}).

\myparagraph{\bnInplace I: Backward pass through \textsc{BN}.} The gradient $\frac{\partial L}{\partial x}=\{\frac{\partial L}{\partial x_1},\dots,\frac{\partial L}{\partial x_m} \}$, which is obtained from the backward pass through the \textsc{BN} layer, can be written as a function of $\hat x=\{\hat x_1,\dots,\hat x_m\}$ and $\frac{\partial L}{\partial y}=\{\frac{\partial L}{\partial y_1},\dots,\frac{\partial L}{\partial y_m} \}$ as 
\[
	\frac{\partial L}{\partial x_i}=\left\{\frac{\partial L}{\partial y_i}-\frac{1}{m}\frac{\partial L}{\partial \gamma}\hat x_i-\frac{1}{m}\frac{\partial L}{\partial\beta}\right\}\frac{\gamma}{\sqrt{\sigma_\set B^2+\epsilon}}\,,
\]
where the gradients of the \textsc{BN} parameters are given by
\[
	\frac{\partial L}{\partial\gamma}=\sum_{i=1}^m\frac{\partial L}{\partial y_i}\hat x_i\,,\quad\frac{\partial L}{\partial \beta}=\sum_{i=1}^m\frac{\partial L}{\partial y_i}\,.
\]
The expression above differs from what is found in the original \textsc{BN} paper~\cite{IofSze15}, but the refactoring was already used in the Caffe~\cite{Jia2014} framework.
It is implemented by $\textsc{BN}_{\gamma,\beta}^*$ in the proposed solutions in Figures~\ref{sfig:mxnet_optim}~and~\ref{sfig:our} and does not depend on $\mu_\set B$. Hence, we store during the forward pass only $\sigma_\set B$ (this dependency was omitted from the diagrams).
Instead, $\textsc{BN}_{\gamma,\beta}$ in Figures~\ref{sfig:standard}~and~\ref{sfig:mxnet}, which depends on $x$, requires the additional recomputation of $\hat x$ from $x$ via Eq.~\eqref{eq:xhat}. Hence, it also requires storing $\mu_\set B$. Our solution is hence memory-wise more efficient than the state-of-the-art from Figure~\ref{sfig:mxnet}.

\myparagraph{Inversion of \boldmath$\pi_{\gamma,\beta}$.} In the configuration of \bnInplace I, the inversion of $\pi_{\gamma,\beta}$ becomes critical if $\gamma=0$ since $\pi^{-1}_{\gamma,\beta}(y)=\frac{y-\beta}{\gamma}$. While we never encountered such a case in practice, one can protect against it by preventing $\gamma$ from getting less than a given tolerance. We can even avoid this problem by simply not considering $\gamma$ a learnable parameter and by fixing it to $1$, in case the activation function is scale covariant (\eg all \textsc{ReLU}-like activations) and when a \textsc{Conv} layer follows. Indeed, it is easy to show that the network retains the exact same capacity in that case, for $\gamma$ can be absorbed into the subsequent \textsc{Conv} layer.

\myparagraph{\bnInplace II: Backward pass through \textsc{BN}.}
We obtain additional memory savings for our solution illustrated in Figure~\ref{sfig:ourII} and as outlined in \S~\ref{ss:overview}. The gradient $\frac{\partial L}{\partial x}$ when written as a function of $y$ instead of $\hat x$ becomes
\[
\frac{\partial L}{\partial x_i}=\left[\frac{\partial L}{\partial y_i}-\frac{1}{\gamma m}\frac{\partial L}{\partial \gamma}y_i-\frac{1}{m}\left(\frac{\partial L}{\partial \beta}-\frac{\beta}{\gamma}\frac{\partial L}{\partial \gamma}\right)\right]\frac{\gamma}{\sqrt{\sigma_\set B^2+\epsilon}}\,.
\]

For the gradients of the \textsc{BN} parameters, $\frac{\partial L}{\partial \beta}$ remains as above but we get 
\[
	\frac{\partial L}{\partial \gamma}=\frac{1}{\gamma}\left[\sum_{j=1}^m\frac{\partial L}{\partial y_j}y_j - \beta\frac{\partial L}{\partial \beta}\right]\,
\]
and we write $\textsc{BN}_{\gamma,\beta}^\dagger$ for the actual backward implementation in Figure~\ref{sfig:ourII}.
Detailed derivations are provided in the Appendix of this paper. 

In summary, both of our optimized main contributions are memory-wise more efficient than the state-of-the-art solution in Figure~\ref{sfig:mxnet} and \bnInplace II is computationally even more efficient than the proposed, optimized checkpointing from Figure~\ref{sfig:mxnet_optim}.


\subsection{Implementation Details}\label{ss:implementation}
We have implemented the proposed \bnInplace I layer in PyTorch, by simply creating a new layer that fuses batch normalization with an (invertible) activation function. In this way we can deal with the computation of $\hat x$ from $z$ internally in the layer, thus keeping the implementation self-contained. We have released code at \url{https://github.com/mapillary/inplace_abn} for easy plug-in replacement of the block \textsc{BN+Act} in modern architectures. The forward and backward implementations are also given as pseudocode in Algorithm~\ref{alg:fwd}~and~\ref{alg:bwd}. In the forward pass, in line 3, we explicitly indicate the buffers that are stored and needed for the backward pass. Any other buffer can be overwritten with in-place computations, \eg $x$, $y$ and $z$ can point to the same memory location. In the backward pass, we recover the stored buffers in line 1 and, again, every computation can be done in-place if the buffer is not needed anymore (\eg $\frac{\partial L}{\partial x}$, $\frac{\partial L}{\partial y}$, $\frac{\partial L}{\partial z}$ can share the same memory location as well as $\hat x$, $y$ and $z$). As opposed to Figure~\ref{fig:overview}, the pseudocode shows also the dependencies on additional, small, buffers like $\sigma_\set B$ and reports the gradients with respect to the \textsc{BN} layer parameters $\gamma$ and $\beta$. Please note the difference during backward computation when applying \bnInplace I or \bnInplace II, respectively.

\setlength{\textfloatsep}{0.4cm}
\setlength{\floatsep}{0.4cm}
\begin{algorithm}[t]
	\begin{algorithmic}[1]
	\Require $x$, $\gamma$, $\beta$
	\State $y, \sigma_\set B\leftarrow\text{BN}_{\gamma,\beta}(x)$ 
	\State $z\leftarrow\phi(y)$ 
	\State save for backward $z$, $\sigma_\set B$
	\State\Return $z$
	\end{algorithmic}
	\caption{\bnInplace Forward}
	\label{alg:fwd}
\end{algorithm}
\begin{algorithm}[t]
	\begin{algorithmic}[1]
	\Require $\frac{\partial L}{\partial z}$, $\gamma$, $\beta$
	\State $z, \sigma_\set B \leftarrow$ saved tensors during forward
	\State $\frac{\partial L}{\partial y}\leftarrow \phi_\text{backward}(z,\frac{\partial L}{\partial z})$
	\State $y\leftarrow\phi^{-1}(z)$ 
	\If{\bnInplace I (see Fig.~\ref{sfig:our})}
	\State $\hat x\leftarrow\pi_{\gamma,\beta}^{-1}(y)$ 
	\State $\frac{\partial L}{\partial x}, \frac{\partial L}{\partial \gamma}, \frac{\partial L}{\partial \beta}\leftarrow \textsc{BN}^*_{\gamma,\beta}(\hat x, \frac{\partial L}{\partial y}, \sigma_{\set B})$
	\ElsIf{\bnInplace II (see Fig.~\ref{sfig:ourII})}
	\State $\frac{\partial L}{\partial x}, \frac{\partial L}{\partial \gamma}, \frac{\partial L}{\partial \beta}\leftarrow \textsc{BN}^\dagger_{\gamma,\beta}(y, \frac{\partial L}{\partial y}, \sigma_{\set B})$ 
    \EndIf	
	\State\Return $\frac{\partial L}{\partial x}$, $\frac{\partial L}{\partial \gamma}$, $\frac{\partial L}{\partial\beta}$
	\end{algorithmic}
	\caption{\bnInplace Backward}
	\label{alg:bwd}
\end{algorithm}
\setlength{\textfloatsep}{0.4cm}
\setlength{\floatsep}{0.4cm}

\section{Experiments}
We assess the effectiveness of our proposed, memory efficient \bnInplace layer for the tasks of image classification and semantic segmentation in \S~\ref{ssec:ExpClass} and~\ref{ssec:ExpSegm}, respectively. Additionally, we provide timing analyses in \S~\ref{ssec:Timings}. Experiments were run and timed on machines comprising four NVIDIA Titan Xp cards (with 12GB of RAM each). Where not otherwise noted, the activation function used in all experiments is \textsc{Leaky ReLU} with slope $a=0.01$.

\subsection{Image Classification}\label{ssec:ExpClass}
We have trained several residual-unit-based models on ImageNet-1k~\cite{Rus+15} to demonstrate the effectiveness of \bnInplace for the task of image classification.
In particular, we focus our attention on two main questions: i) whether using an invertible activation function (\ie \textsc{Leaky ReLU} in our experiments) impacts on the performance of the models, and ii) how the memory savings obtained with our method can be exploited to improve classification accuracy.
Our results are summarized in Table~\ref{tab:imagenet} and described in this subsection.

\myparagraph{\texttt{ResNeXt-101}/\texttt{ResNeXt-152}.}
This is a variant of the original ResNet~\cite{He2015b} architecture in which the bottleneck residual units are replaced with a multi-branch version.
In practice, this equates to ``grouping'' the $3\times 3$ convolutions of each residual unit.
The number of groups, \ie parallel branches, is known as \textit{cardinality} and is an additional hyperparameter to be set.
We follow the best performing design guidelines described in~\cite{Xie2016} and train models with cardinality 64, while considering the 101- and 152-layers configurations.
During training, we proportionally scale input images so that their smallest side equals $256$ pixels, before randomly taking $224\times 224$ crops.
Additionally, we perform per-channel mean and variance normalization and color augmentation as described in~\cite{Xie2016}.
We train using stochastic gradient descent (SGD) with Nesterov updates, initial learning rate $0.1$, weight decay $10^{-4}$ and momentum $0.9$.
The training is run for a total of $90$ epochs, reducing the learning rate every $30$ epochs by a factor $10$.

\myparagraph{\texttt {WideResNet-38.}}
This is another modern architecture built by stacking residual units.
Compared to the original ResNet, WideResNet trades depth for \emph{width}, \ie it uses units with an increased number of feature channels while reducing the total number of stacked units.
For training, we use the same setup and hyperparameters as for ResNeXt, with one exception: following~\cite{Wu2016c} we train for $90$ epochs, linearly decreasing the learning rate from $0.1$ to $10^{-6}$. 

\myparagraph{Discussion of results.}
In our experiments we also compared the validation accuracy obtained when replacing \textsc{ReLU} with \textsc{Leaky ReLU} in a \texttt{ResNeXt-101} trained with \textsc{ReLU}. We also considered the opposite case, replacing \textsc{Leaky ReLU} with \textsc{ReLU} in a \textsc{Leaky ReLU}-trained network (see Table~\ref{tab:imagenetReluLRelu}). Our results are in line with~\cite{Xu+15}, and never differ by more than a single point per training except for the $320^2$ center crop evaluation top-1 results, probably also due to non-deterministic training behaviour.

\begin{table*}[t]
\centering
\small
\begin{tabular}{lcccccccc}
  \toprule
    \multirow{2}{*}{Network} & \multicolumn{2}{c}{activation} & \multicolumn{2}{c}{$224^2$ center} & \multicolumn{2}{c}{$224^2$ 10-crops} & \multicolumn{2}{c}{$320^2$ center} \\
  \cmidrule{4-9}
    & training & validation & top-1 & top-5 & top-1 & top-5 & top-1 & top-5 \\
  \midrule
    \texttt{ResNeXt-101} & \textsc{ReLU} & \textsc{ReLU} & 77.74 & 93.86 & 79.21 & 94.67 & 79.17 & 94.67 \\
    \texttt{ResNeXt-101} & \textsc{ReLU} & \textsc{Leaky ReLU} & 76.88 & 93.42 & 78.74 & 94.46 & 78.37 & 94.25 \\
  \midrule
    \texttt{ResNeXt-101} & \textsc{Leaky ReLU} & \textsc{Leaky ReLU} & 77.04 & 93.50 & 78.72 & 94.47 & 77.92 & 94.28 \\
    \texttt{ResNeXt-101} & \textsc{Leaky ReLU} & \textsc{ReLU} & 76.81 & 93.53 & 78.46 & 94.38 & 77.84 & 94.20 \\
  \bottomrule
\end{tabular}
  \vspace{1em}
  \caption{Imagenet validation set results using \texttt{ResNeXt-101} and \textsc{ReLU}/\textsc{Leaky ReLU} exchanged activation functions during training and validation.}
  \label{tab:imagenetReluLRelu}
\end{table*}

Our results may slightly differ from what was reported in the original papers, as our training protocol does not exactly match the one in~\cite{Xie2016} (\eg data augmentation regarding scale and aspect ratio settings, learning rate schedule, \etc) or due to changes in reference implementations.\footnote{See note in \url{https://github.com/itijyou/ademxapp} mentioning subtle differences in implementation, \eg, different cropping strategies, interpolation methods, and padding strategies.}
\begin{table*}[t]
\centering
\small
\begin{tabular}{lccccccc}
  \toprule
    \multirow{2}{*}{Network} & & \multicolumn{2}{c}{$224^2$ center} & \multicolumn{2}{c}{$224^2$ 10-crops} & \multicolumn{2}{c}{$320^2$ center} \\
  \cmidrule{3-8}
    & batch size & top-1 & top-5 & top-1 & top-5 & top-1 & top-5 \\
  \midrule
    \texttt{ResNeXt-101}, \bnStd & 256 & 77.04 & 93.50 & 78.72 & 94.47 & 77.92 & 94.28 \\
  \midrule
    \texttt{ResNeXt-101}, \bnInplace & 512 & 78.08 & 93.79 & 79.52 & 94.66 & 79.38 & 94.67 \\
    \texttt{ResNeXt-152}, \bnInplace & 256 & 78.28 & 94.04 & 79.73 & 94.82 & 79.56 & 94.67 \\
    \texttt{WideResNet-38}, \bnInplace & 256 & 79.72 & 94.78 & 81.03 & 95.43 & 80.69 & 95.27 \\
  \midrule
    \texttt{ResNeXt-101}, \bnInplaceSync & 256 & 77.70 & 93.78 & 79.18 & 94.60 & 78.98 & 94.56 \\
  \bottomrule
\end{tabular}
  \vspace{1em}
  \caption{Imagenet validation set results using different architectures and training batch sizes.}
  \label{tab:imagenet}
\end{table*}
Next, we focus on how to better exploit the memory savings due to our proposed \bnInplace for improving classification accuracy.
As a baseline, we train \texttt{ResNeXt-101} with standard Batch Normalization and the maximum batch size that fits in GPU memory, \ie 256 images per batch.
Then, we consider two different options: i) using the extra memory to fit more images per training batch while fixing the network architecture, or ii) fixing the batch size while training a larger network.
For option i) we double the batch size to 512 (\texttt{ResNeXt-101}, \bnInplace, 512 in Table~\ref{tab:imagenet}), while for option ii) we train \texttt{ResNeXt-152} and \texttt{WideResNet-38}.
Note that neither \texttt{ResNeXt-152} nor \texttt{WideResNet-38} would fit in memory when using 256 images per training batch and when using standard \textsc{BN}.
As it is clear from the table, both i) and ii) result in a noticeable performance increase.
Interestingly, training \texttt{ResNeXt-101} with an increased batch size results in similar accuracy to the deeper (and computationally more expensive) \texttt{ResNeXt-152} model.
As an additional reference, we train \texttt{ResNeXt-101} with synchronized Batch Normalization (\bnInplaceSync), which can be seen as a ``virtual'' increase of batch size applied to the computation of \textsc{BN} statistics.
In this case we only observe small accuracy improvements when compared to the baseline model.
For the future, we plan to conduct further experiments with deeper variants of DenseNet~\cite{Huang2017}, and investigate effects of \bnInplace on Squeeze~\&~Excitation networks~\cite{Hu2017} or deformable convolutional networks~\cite{Dai+17}.



\subsection{Semantic Segmentation}\label{ssec:ExpSegm}
The goal of semantic segmentation is to assign categorical labels to each pixel in an image. State-of-the-art segmentations are typically obtained by combining classification models pretrained on ImageNet (typically referred to as \textit{body}) with segmentation-specific \textit{head} architectures and jointly fine-tuning them on suitable, (densely) annotated training data like Cityscapes~\cite{Cordts2016}, COCO-Stuff~\cite{Caesar2016}, ADE20K~\cite{Zho+16} or Mapillary Vistas~\cite{Neuhold2017}. 

\myparagraph{Datasets used for Evaluation.}
We report results on Cityscapes~\cite{Cordts2016}, COCO-Stuff~\cite{Caesar2016} and Mapillary Vistas~\cite{Neuhold2017}, since these datasets have complementary properties in terms of image content, size, number of class labels and annotation quality. Cityscapes shows street-level images captured in central Europe and comprises a total of 5k densely annotated images (19 object categories + 1 \textit{void} class, all images sized 2048$\times$1024), split into 2975/500/1525 images for training, validation and test, respectively. While there exist additional 20k images with so-called \textit{coarse} annotations, we learn only from the high-quality (fine) annotations in the \textit{training} set and test on the corresponding \textit{validation} set (for which ground truth is publicly available).
We also show results on COCO-Stuff, which provides \textit{stuff}-class annotations for the well-known MS COCO dataset~\cite{LinMSCOCO2014}. This dataset comprises 65k COCO images (with 40k for training, 5k for validation, 5k for test-dev and 15k as challenge test set) with annotations for 91 stuff classes and 1 \textit{void} class. Images are smaller than in Cityscapes and with varying sizes, and the provided semantic annotations are based on superpixel segmentations, consequently suffering from considerable mislabelings. 
Finally, we also report results on Mapillary Vistas (research edition), a novel and large-scale street-level image dataset comprising 25k densely annotation images (65 object categories + 1 \textit{void} class, images have varying aspect ratios and sizes up to 22 Megapixel), split into 18k/2k/5k images for training, validation and test, respectively. Similar to the aforementioned datasets, we train on training data and test on validation data.
\begin{table*}[t]
\centering
\resizebox{\textwidth}{!}{
\begin{tabular}{llclclclc}
  \toprule 
    \multirow{2}{*}{\textsc{BatchNorm}} & \multicolumn{4}{c}{\texttt{ResNeXt-101}} & \multicolumn{4}{c}{\texttt{WideResNet-38}} \\
    \cmidrule{2-9}
     & \multicolumn{2}{c}{\gc Cityscapes} & \multicolumn{2}{c}{COCO-Stuff} & \multicolumn{2}{c}{\gc Cityscapes} & \multicolumn{2}{c}{COCO-Stuff} \\
    \bnStd + \textsc{Leaky ReLU} & \gc $16\times 512^2$ & \gc 74.42 & $16\times 480^2$ & 20.30 &\gc $20\times 512^2$ &\gc 75.82  &$20\times 496^2$ & 22.44\\
    \midrule
    \bnInplace, \fixedcrop & \gc $28\times 512^2$ {\footnotesize \textbf{[+75\%]}}      & \gc 75.80  & $24\times 480^2$ {\footnotesize \textbf{[+50\%]}} & 22.63 &\gc $28\times 512^2$ {\footnotesize \textbf{[+40\%]}} &\gc 77.75  &$28\times 496^2$ {\footnotesize \textbf{[+40\%]}} & 22.96\\
    \bnInplace, \fixedbatch & \gc $16\times 672^2$ {\footnotesize \textbf{[+72\%]}}     & \gc 77.04 & $16\times 600^2$ {\footnotesize \textbf{[+56\%]}} & 23.35 &\gc $20\times 640^2$ {\footnotesize \textbf{[+56\%]}} &\gc \textbf{78.31}  &$20\times 576^2$ {\footnotesize \textbf{[+35\%]}} & 24.10\\
    \bnInplaceSync, \fixedbatch & \gc $16\times 672^2$ {\footnotesize \textbf{[+72\%]}} & \gc \textbf{77.58} 
      & $16\times 600^2$ {\footnotesize \textbf{[+56\%]}} & \textbf{24.91} &\gc $20\times 640^2$ {\footnotesize \textbf{[+56\%]}} &\gc 78.06  &$20\times 576^2$ {\footnotesize \textbf{[+35\%]}} & \textbf{25.11}\\
\bottomrule
\end{tabular}
}
\vspace{2pt}
   \caption{Validation data results (single scale test) for semantic segmentation experiments on Cityscapes and COCO-Stuff, using \texttt{ResNeXt-101} and \texttt{WideResNet-38} network bodies and different batch normalization settings (see text). All result numbers in [\%].}
\label{tab:SegmentationResults}
\end{table*}

\myparagraph{Segmentation approach.}
We chose to adopt the recently introduced DeepLabV3~\cite{Chen2017} segmentation approach as head, and evaluate its performance with body networks from \S~\ref{ssec:ExpClass}. DeepLabV3 is exploiting atrous (dilated) convolutions in a cascaded way for capturing contextual information, together with crop-level features encoding global context (close in spirit to PSPNet's~\cite{zhao2016pspnet} global feature). We follow the parameter choices suggested in~\cite{Chen2017}, assembling the head as 4 parallel \textsc{Conv} blocks with $256$ output channels each and dilation rates $(1,12,24,36)$ (with x8 downsampled crop sizes from the body) and kernel sizes $(1^2,3^2,3^2,3^2)$, respectively. The global $1\times1$ features are computed in a channel-specific way and \textsc{Conv}ed into $256$ additional channels. Each output block is followed by BatchNorm before all $1280$ features are stacked and reduced by another \textsc{Conv+BN+Act} block (into 256 features) and finally \textsc{Conv}ed to the number of target classes. We exploit our proposed \bnInplace strategy also in the head architecture. Finally, we apply bilinear upsampling to the logits to obtain the original input crop resolution before computing the loss using an \textit{online bootstrapping} strategy as described in~\cite{Wu2016, RotNeuKon17cvpr} (setting $p=1.0$ and $m=25\%$). We did not apply hybrid dilated convolutions~\cite{Wang2017} nor added an auxiliary loss as proposed in~\cite{zhao2016pspnet}. Training data is sampled in a uniform way (by shuffling the database in each epoch) and all Cityscapes experiments are run for 360 epochs using an initial learning rate of $2.5\times 10^{-3}$ and polynomial learning rate decay $(1-\frac{iter}{max\_iter})^{0.9}$, following~\cite{Chen2017}. COCO-Stuff experiments were trained only for 30 epochs, which however approximately matches the number of iterations on Cityscapes due to the considerably larger dataset size. For optimization, we use stochastic gradient descent with momentum $0.9$ and weight decay $10^{-4}$. For training data augmentation, we apply random horizontal flipping (with prob. $0.5$) and random scaling selected from $0.7$ - $2.0$ before cropping the actual patches. 

\myparagraph{Discussion of Results.}
In Table~\ref{tab:SegmentationResults}, we provide results on validation data for Cityscapes and COCO-Stuff under different \textsc{BN} layer configurations. We distinguish between standard \textsc{BN} layers~\cite{IofSze15} (coined \bnStd) and our proposed variants using in-place, activated \textsc{BN} (\bnInplace) as well as its gradient-synchronized version \bnInplaceSync. All experiments are based on \textsc{Leaky ReLU} activations. Trainings were conducted in a way to maximize GPU memory utilization by \textit{i)} fixing the training crop size and therefore pushing the amount of crops per minibatch to the limit (denoted as \fixedcrop) or \textit{ii)} fixing the number of crops per minibatch and maximizing the training crop resolutions (\fixedbatch). Experiments are conducted for \texttt{ResNeXt-101} and \texttt{WideResNet-38} network bodies, where the latter seems preferable for segmentation tasks. Both body networks were solely trained on ImageNet-1k. All results derive from single-scale testing without horizontal image flipping, deliberately avoiding dilution of results by well-known bells and whistles.
Table~\ref{tab:SegmentationResults} shows the positive effects of applying more training data (in terms of both, \#training crops per minibatch and input crop resolutions) on the validation results. The increase of data (\wrt pixels/minibatch) we can put in GPU memory, relative to the baseline (top row) is reported in square brackets. We observe that higher input resolution is in general even more beneficial than adding more crops to the batch. For the sake of direct comparability we left the learning rates unchanged, but there might be better hyper-parameters for our variants of \bnInplace and \bnInplaceSync. In essence, our results closely approach reported numbers for Cityscapes in \eg~\cite{Gadde2017,zhao2016pspnet}, which however include more training data in the body model training or are based on already trained models like DeepLabV2~\cite{Chen2016}. For COCO-Stuff, results are not directly comparable to the original publication~\cite{Caesar2016} since they used a lower-performing \texttt{VGG-16}~\cite{Simonyan2014} network for the body. However, all experiments show significant improvements \wrt the baseline method in the first line of Table~\ref{tab:SegmentationResults}.

\myparagraph{Optimizing Settings for Cityscapes and Vistas Datasets.}
Here we show further experiments when tuning settings like \#training crops and crop sizes in favor of our method (as opposed to maintaining comparability with baselines above). 
First, we report Cityscapes results in Table~\ref{tab:moreresults} when fine-tuning our \bnInplace \texttt{ResNeXt-152} ImageNet model from \S~\ref{ssec:ExpClass}, using 12 crops of size $680\times680$ per minibatch (using the gradient-synchronized variant \bnInplaceSync). Going deeper with the body results in a validation set score of \textbf{78.49\%}, improving over the score of 77.58\% obtained by \texttt{ResNeXt-101}, \bnInplaceSync. We provide additional results using \texttt{WideResNet-38} \bnInplaceSync-based settings, where we trained the model with \textit{i)} 16 crops at $712\times712$ yielding 79.02\% and \textit{ii)} 12 crops at $872\times872$, resulting in \textbf{79.16\%}. As can be seen, the combination of \bnInplaceSync with larger crop sizes improves by $\approx0.9\%$ over the best performing setting in Table~\ref{tab:SegmentationResults} (\bnInplace with 20 crops at $640\times640$). We also list a non-gradient synchronized experiment (as this gave the highest score on Cityscapes before), where an increase of the crop size yields to minor improvements, climbing from 78.31\% to 78.45\%. 

Finally, we have run another experiment with 12 crops at $872\times872$ where we however used a different training data sampling strategy. Instead of just randomly perturbing the dataset and taking training crops from random positions, we compiled the minibatches per epoch in a way to show all classes approximately uniformly (thus following an oversampling strategy for underrepresented categories). 
In practice, we tracked object class presence for all images and eventually class-uniformly sampled from eligible image candidates, making sure to take training crops from areas containing the class of interest. Applying this sampling strategy coined \textsc{Class-Uniform sampling} yields \textbf{79.40\%}, which matches the highest reported score on Cityscapes validation data reported in~\cite{Gadde2017}, without however using additional training data.\\
Next, we provide results for the Mapillary Vistas dataset, using hyperparameter settings inspired by our highest scoring configuration for Cityscapes. Vistas is considerably larger than Cityscapes (in terms of \#classes, \#images and image resolution), so running an exhaustive amount of experiments is prohibitive in terms of training time. Due to the increase of object classes (19 for Cityscapes and 65 for Vistas), we used minibatches of 12 crops at $776\times776$ (with \bnInplaceSync), increased the initial learning rate to $3.5\times 10^{-3}$ and trained for 90 epochs. This setting leads to the highest reported single-scale score of \textbf{53.12\%} on validation data so far, significantly outperforming the LSUN 2017 segmentation winner's single-scale approach~\cite{LSUNSeg17} of 51.59\%. As also listed in Table~\ref{tab:moreresults}, their approach additionally used hybrid dilated convolutions~\cite{Wang2017}, applied an inverse frequency weighting for correcting training data class imbalance as well as pretrained on Cityscapes.

\begin{table*}[t]
\centering
\small
\begin{tabular}{lcccc}
  \toprule 
      & \multicolumn{2}{c}{\texttt{ResNeXt-152}} & \multicolumn{2}{c}{\texttt{WideResNet-38}} \\
	\midrule
    \textbf{Cityscapes} & & & & \\
    \bnInplaceSync & $12\times 680^2$  & 78.49 & \multicolumn{2}{c}{--}\\
    \bnInplace & \multicolumn{2}{c}{--} & $16\times 712^2$ & 78.45 \\
    \bnInplaceSync & \multicolumn{2}{c}{--} & $16\times 712^2$ & 79.02 \\
    \bnInplaceSync & \multicolumn{2}{c}{--} & $12\times 872^2$ & 79.16 \\ 
    \bnInplaceSync + \textsc{Class-Uniform sampling} & \multicolumn{2}{c}{--} & $12\times 872^2$ & \textbf{79.40} \\
    \midrule
    \rowcolor{gray!25} \textbf{Mapillary Vistas} & & & & \\
    \rowcolor{gray!25} \bnInplaceSync + \textsc{Class-Uniform sampling} & \multicolumn{2}{c}{--} & $12\times 776^2$ & \textbf{53.12} \\
    \rowcolor{gray!25}\midrule        
    \rowcolor{gray!25}LSUN 2017 winner~\cite{LSUNSeg17} (based on PSPNet) & \multicolumn{4}{c}{\texttt{ResNet-101}} \\
    \rowcolor{gray!25}PSPNet + auxiliary loss & \multicolumn{2}{c}{$16\times 713^2$} & \multicolumn{2}{c}{49.76}\\
    \rowcolor{gray!25}$\quad$+ Hybrid dilated convolutions~\cite{Wang2017} & \multicolumn{2}{c}{$16\times 713^2$} & \multicolumn{2}{c}{50.28}\\
    \rowcolor{gray!25}$\quad$+ Inverse frequency label reweighting & \multicolumn{2}{c}{$16\times 713^2$} & \multicolumn{2}{c}{51.50}\\
    \rowcolor{gray!25}$\quad$+ Cityscapes pretraining & \multicolumn{2}{c}{$16\times 713^2$} & \multicolumn{2}{c}{51.59}\\

  \bottomrule
\end{tabular}
\vspace{5pt}
   \caption{Validation data results (single scale test, no horizontal flipping) for semantic segmentation experiments on Cityscapes and Vistas, using \texttt{ResNeXt-152} and \texttt{WideResNet-38} bodies with different settings for \#crops per minibatch and crop sizes. All results in [\%].}
\label{tab:moreresults}
\end{table*}

\subsection{Timing analyses}\label{ssec:Timings}
Besides the discussed memory improvements and their impact on computer vision applications, we also provide actual runtime comparisons and analyses for the \bnInplace I setting shown in~\ref{sfig:our}, as this is the implementation we made publicly available\footnote{\url{https://github.com/mapillary/inplace_abn}}.
\begin{figure}[b]
  \centering
  \includegraphics[width=\columnwidth]{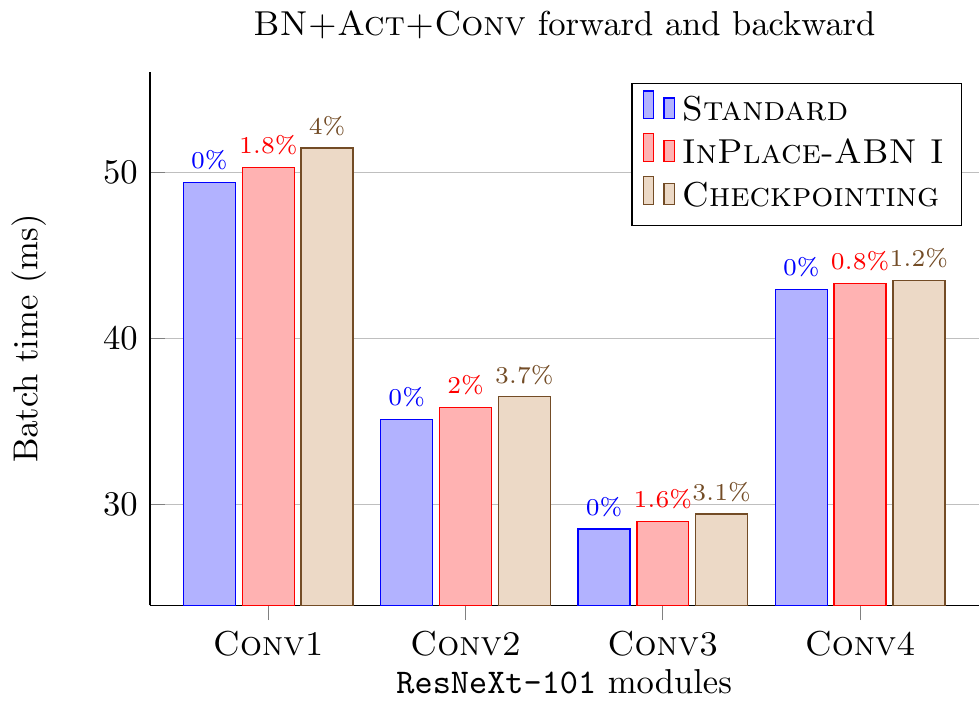}
  \caption{Computation time required for a forward and backward pass through basic \textsc{BN+Act+Conv} blocks
  from \texttt{ResNeXt-101}, using different \textsc{BN} strategies.}
  \label{fig:times}
\end{figure}
Isolating a single \textsc{BN+Act+Conv} block, we evaluate the computational times required for a forward and backward pass over it (Figure~\ref{fig:times}). We compare the conventional approach of serially executing layers and storing intermediate results (\textsc{Standard}), our proposed \bnInplace I and the \textsc{Checkpointing} approach. In order to obtain fair timing comparisons, we re-implemented the checkpointing idea in PyTorch. 
The results are obtained by running all operations over a batch comprising 32-images and setting the meta-parameters (number of feature channels, spatial dimensions) to those encountered in the four modules of \texttt{ResNeXt-101}, denoted as \textsc{Conv1}-\textsc{Conv4}. The actual runtimes were averaged over 200 iterations. 

We observe consistent speed advantages in favor of our method when comparing against \textsc{Checkpointing}, with the actual percentage difference depending on block's meta-parameters. As we can see, \bnInplace induces computation time increase between $0.8-2\%$ over \textsc{Standard} while \textsc{Checkpointing} is almost doubling our overheads.

\section{Conclusions}
In this work we have presented \bnInplace, which is a novel, computationally efficient fusion of batch normalization and activation layers, targeting memory-optimization for modern deep neural networks during training time. We reconstruct necessary quantities for the backward pass by inverting the forward computation from the storage buffer, and manage to free up almost 50\% of the memory needed in conventional \textsc{BN+Act} implementations at little additional computational costs. In contrast to state-of-the-art checkpointing attempts, our method is reconstructing discarded buffers \textit{backwards} during the backward pass, thus allowing us to encapsulate \textsc{BN+Act} as self-contained layer, which is easy to implement and deploy in virtually all modern deep learning frameworks. We have validated our approach with experiments for image classification on ImageNet-1k and semantic segmentation on Cityscapes, COCO-Stuff and Mapillary Vistas. Our obtained networks have performed consistently and considerably better when trained with larger batch sizes (or training crop sizes), leading to a new high-score on the challenging Mapillary Vistas dataset in a single-scale, single-model inference setting. 
In future works, we will investigate the consequences of our approach for problems like object detection, instance-specific segmentation and learning in 3D. Derivations for gradient computation are provided in the Appendix.

\myparagraph{Acknowledgements.} We acknowledge financial support from project \textit{DIGIMAP}, funded under grant \#860375 by the Austrian Research Promotion Agency (FFG).

\begin{widetext}
\section*{Appendix -- Derivation of Gradient \boldmath $\frac{\partial L}{\partial x}$}\label{sec:Appendix}
We follow the gradient derivations as provided in the original batch normalization paper~\cite{IofSze15} and rewrite them as a function of $\hat x$, starting with generally required derivatives for \bnInplace and particular ones of \bnInplace I.
\begin{align*}
	&\frac{\partial y_j}{\partial \gamma} = \hat x_j\,,&&\frac{\partial y_j}{\partial \beta} = 1\,,&&\frac{\partial y_j}{\partial \hat x_j}=\gamma\,,\\
	&\frac{\partial L}{\partial \gamma}=\sum_{j=1}^m \frac{\partial L}{\partial y_j}\frac{\partial y_j}{\partial \gamma}=\sum_{j=1}^m\frac{\partial L}{\partial y_j}\hat x_j\,,&
	&\frac{\partial L}{\partial \beta}=\sum_{j=1}^m \frac{\partial L}{\partial y_j}\frac{\partial y_j}{\partial \beta}=\sum_{j=1}^m\frac{\partial L}{\partial y_j}\,,&
	&\frac{\partial  L}{\partial \hat x_j}=\frac{\partial L}{\partial y_j}\frac{\partial y_j}{\partial \hat x_j}=\frac{\partial L}{\partial y_j}\gamma\,,
\end{align*}
\begin{align*}
	&\frac{\partial \hat x_j}{\partial \sigma_\set B^2}=-\frac{1}{2(\sigma_\set B^2+\epsilon)}\frac{x_j-\mu_\set B}{\sqrt{\sigma_\set B^2+\epsilon}}=-\frac{\hat x_j}{2(\sigma_\set B^2+\epsilon)}\,,&
	&\frac{\partial \hat x_j}{\partial \mu_\set B}=-\frac{1}{\sqrt{\sigma_\set B^2+\epsilon}}\,,
\end{align*}
\begin{align*}
	\frac{\partial  L}{\partial \sigma^2_{\set B}}&=\sum_{j=1}^m\frac{\partial L}{\partial \hat x_j}\frac{\partial\hat x_j}{\partial \sigma_\set B^2}=-\frac{\gamma}{2(\sigma_\set B^2+\epsilon)}\sum_{j=1}^m\frac{\partial L}{\partial y_j}\hat x_j=-\frac{\gamma}{2(\sigma_\set B^2+\epsilon)}\frac{\partial L}{\partial \gamma}\,,\\
	\frac{\partial  L}{\partial\mu_{\set B}}&=\sum_{j=1}^m\frac{\partial L}{\partial \hat x_j}\frac{\partial \hat x_j}{\partial \mu_\set B}=-\frac{\gamma}{\sqrt{\sigma_\set B^2+\epsilon}}\sum_{j=1}^m\frac{\partial L}{\partial y_j}=-\frac{\gamma}{\sqrt{\sigma_\set B^2+\epsilon}}\frac{\partial L}{\partial \beta}\,,
\end{align*}

\begin{align*}
	&\frac{\partial \sigma^2_\set B}{\partial x_i}=\frac{2(x_i-\mu_\set B)}{m}\,,&
	&\frac{\partial \mu_\set B}{\partial x_i}=\frac{1}{m}\,,&
	&\frac{\partial \hat x_i}{\partial x_i}=\frac{1}{\sqrt{\sigma_\set B^2+\epsilon}}\,,
\end{align*}
\[
	\frac{\partial L}{\partial x_i}=\frac{\partial L}{\partial \hat x_i}\frac{\partial\hat x_i}{\partial x_i}+\frac{\partial L}{\partial \sigma^2_\set B}\frac{\partial \sigma^2_\set B}{\partial x_i}+\frac{\partial L}{\partial \mu_\set B}\frac{\partial \mu_\set B}{\partial x_i}=\left(\frac{\partial L}{\partial y_i}-\frac{1}{m}\frac{\partial L}{\partial \gamma}\hat x_i-\frac{1}{m}\frac{\partial L}{\partial \beta}\right)\frac{\gamma}{\sqrt{\sigma_\set B^2+\epsilon}}\,.
\]

For \bnInplace II, we write gradients $\frac{\partial L}{\partial \gamma}$ and $\frac{\partial L}{\partial x}$ as functions of $y$ instead of $\hat x$ in the following way:
\[
	\frac{\partial L}{\partial \gamma}=\sum_{j=1}^m\frac{\partial L}{\partial y_j}\hat x_j=\sum_{j=1}^m\frac{\partial L}{\partial y_j}\frac{y_j-\beta}{\gamma}=\frac{1}{\gamma}\sum_{j=1}^m\frac{\partial L}{\partial y_j}y_j - \frac{\beta}{\gamma}\sum_{j=1}^m\frac{\partial L}{\partial y_j}=\frac{1}{\gamma}\left[\sum_{j=1}^m\frac{\partial L}{\partial y_j}y_j - \beta\frac{\partial L}{\partial \beta}\right]\,,
\]
\begin{align*}
\frac{\partial L}{\partial x_i}&=\left(\frac{\partial L}{\partial y_i}-\frac{1}{m}\frac{\partial L}{\partial \gamma}\hat x_i-\frac{1}{m}\frac{\partial L}{\partial \beta}\right)\frac{\gamma}{\sqrt{\sigma_\set B^2+\epsilon}}\\
&=\left(\frac{\partial L}{\partial y_i}-\frac{1}{m}\frac{\partial L}{\partial \gamma}\frac{y_i-\beta}{\gamma}-\frac{1}{m}\frac{\partial L}{\partial \beta}\right)\frac{\gamma}{\sqrt{\sigma_\set B^2+\epsilon}}\\
&=\left[\frac{\partial L}{\partial y_i}-\frac{1}{\gamma m}\frac{\partial L}{\partial \gamma}y_i-\frac{1}{m}\left(\frac{\partial L}{\partial \beta}-\frac{\beta}{\gamma}\frac{\partial L}{\partial \gamma}\right)\right]\frac{\gamma}{\sqrt{\sigma_\set B^2+\epsilon}}\,.
\end{align*}

\end{widetext}
{\small
\bibliographystyle{ieee}
\bibliography{../bib/GeneralRefs}

\begin{thebibliography}{10}\itemsep=-1pt

\bibitem{Caesar2016}
H.~Caesar, J.~R.~R. Uijlings, and V.~Ferrari.
\newblock {COCO-S}tuff: Thing and stuff classes in context.
\newblock {\em CoRR}, abs/1612.03716, 2016.

\bibitem{Chen2016}
L.~Chen, G.~Papandreou, I.~Kokkinos, K.~Murphy, and A.~L. Yuille.
\newblock Deeplab: Semantic image segmentation with deep convolutional nets,
  atrous convolution, and fully connected {CRF}s.
\newblock {\em CoRR}, abs/1606.00915, 2016.

\bibitem{Chen2017}
L.~Chen, G.~Papandreou, F.~Schroff, and H.~Adam.
\newblock Rethinking atrous convolution for semantic image segmentation.
\newblock {\em CoRR}, abs/1706.05587, 2017.

\bibitem{Chen+16}
T.~Chen, B.~Xu, C.~Zhang, and C.~Guestrin.
\newblock Training deep nets with sublinear memory cost.
\newblock {\em CoRR}, abs/1604.06174, 2016.

\bibitem{Cordts2016}
M.~Cordts, M.~Omran, S.~Ramos, T.~Rehfeld, M.~Enzweiler, R.~Benenson,
  U.~Franke, S.~Roth, and B.~Schiele.
\newblock The {C}ityscapes dataset for semantic urban scene understanding.
\newblock In {\em (CVPR)}, 2016.

\bibitem{Courbariaux2015}
M.~Courbariaux, Y.~Bengio, and J.-P. David.
\newblock Binaryconnect: Training deep neural networks with binary weights
  during propagations.
\newblock In {\em (NIPS)}. 2015.

\bibitem{Dai+17}
J.~Dai, H.~Qi, Y.~Xiong, Y.~Li, G.~Zhang, H.~Hu, and Y.~Wei.
\newblock Deformable convolutional networks.
\newblock {\em CoRR}, abs/1703.06211, 2017.

\bibitem{Gadde2017}
R.~Gadde, V.~Jampani, and P.~V. Gehler.
\newblock Semantic video cnns through representation warping.
\newblock {\em CoRR}, abs/1708.03088, 2017.

\bibitem{Gomez2017}
A.~N. Gomez, M.~Ren, R.~Urtasun, and R.~B. Grosse.
\newblock The reversible residual network: Backpropagation without storing
  activations.
\newblock In {\em (NIPS)}, December 2017.

\bibitem{Gruslys2016}
A.~Gruslys, R.~Munos, I.~Danihelka, M.~Lanctot, and A.~Graves.
\newblock Memory-efficient backpropagation through time.
\newblock In {\em (NIPS)}, 2016.

\bibitem{He2015b}
K.~He, X.~Zhang, S.~Ren, and J.~Sun.
\newblock Deep residual learning for image recognition.
\newblock {\em CoRR}, abs/1512.03385, 2015.

\bibitem{He+16}
K.~He, X.~Zhang, S.~Ren, and J.~Sun.
\newblock Identity mappings in deep residual networks.
\newblock {\em CoRR}, abs/1603.05027, 2016.

\bibitem{Hu2017}
J.~Hu, L.~Shen, and G.~Sun.
\newblock Squeeze-and-excitation networks.
\newblock {\em CoRR}, abs/1709.01507, 2017.

\bibitem{Huang2017}
G.~Huang, Z.~Liu, L.~van~der Maaten, and K.~Q. Weinberger.
\newblock Densely connected convolutional networks.
\newblock In {\em (CVPR)}, July 2017.

\bibitem{Hubara2016a}
I.~Hubara, M.~Courbariaux, D.~Soudry, R.~El-Yaniv, and Y.~Bengio.
\newblock Binarized neural networks.
\newblock In {\em (NIPS)}. 2016.

\bibitem{Hubara2016b}
I.~Hubara, M.~Courbariaux, D.~Soudry, R.~El{-}Yaniv, and Y.~Bengio.
\newblock Quantized neural networks: Training neural networks with low
  precision weights and activations.
\newblock {\em CoRR}, abs/1609.07061, 2016.

\bibitem{IofSze15}
S.~Ioffe and C.~Szegedy.
\newblock Batch normalization: Accelerating deep network training by reducing
  internal covariate shift.
\newblock {\em CoRR}, abs/1502.03167, 2015.

\bibitem{Jia2014}
Y.~Jia, E.~Shelhamer, J.~Donahue, S.~Karayev, J.~Long, R.~Girshick,
  S.~Guadarrama, and T.~Darrell.
\newblock Caffe: Convolutional architecture for fast feature embedding.
\newblock {\em arXiv preprint arXiv:1408.5093}, 2014.

\bibitem{LinMSCOCO2014}
T.~Lin, M.~Maire, S.~J. Belongie, L.~D. Bourdev, R.~B. Girshick, J.~Hays,
  P.~Perona, D.~Ramanan, P.~Doll{\'{a}}r, and C.~L. Zitnick.
\newblock Microsoft {COCO:} {C}ommon objects in context.
\newblock {\em CoRR}, abs/1405.0312, 2014.

\bibitem{MaaHanNg13}
A.~L. Maas, A.~Y. Hannun, and A.~Y. Ng.
\newblock Rectifier nonlinearities improve neural network acoustic models.
\newblock In {\em in ICML Workshop on Deep Learning for Audio, Speech and
  Language Processing}, 2013.

\bibitem{Martens2012}
J.~Martens and I.~Sutskever.
\newblock {\em Training Deep and Recurrent Networks with Hessian-Free
  Optimization}, pages 479--535.
\newblock Springer Berlin Heidelberg, 2012.

\bibitem{Micikevicius2017}
P.~Micikevicius, S.~Narang, J.~Alben, G.~F. Diamos, E.~Elsen, D.~Garcia,
  B.~Ginsburg, M.~Houston, O.~Kuchaiev, G.~Venkatesh, and H.~Wu.
\newblock Mixed precision training.
\newblock {\em CoRR}, abs/1710.03740, 2017.

\bibitem{Neuhold2017}
G.~Neuhold, T.~Ollmann, S.~Rota~Bul\`o, and P.~Kontschieder.
\newblock The mapillary vistas dataset for semantic understanding of street
  scenes.
\newblock In {\em (ICCV)}, October 2017.

\bibitem{Pleiss+17}
G.~Pleiss, D.~Chen, G.~Huang, T.~Li, L.~van~der Maaten, and K.~Q. Weinberger.
\newblock Memory-efficient implementation of densenets.
\newblock {\em CoRR}, abs/1707.06990, 2017.

\bibitem{RotNeuKon17cvpr}
S.~Rota~Bul\`o, G.~Neuhold, and P.~Kontschieder.
\newblock Loss max-pooling for semantic image segmentation.
\newblock In {\em (CVPR)}, July 2017.

\bibitem{Rus+15}
O.~Russakovsky, J.~Deng, H.~Su, J.~Krause, S.~Satheesh, S.~Ma, Z.~Huang,
  A.~Karphathy, A.~Khosla, M.~Bernstein, A.~C. Berg, and L.~Fei-Fei.
\newblock Imagenet large scale visual recognition challenge.
\newblock {\em (IJCV)}, 2015.

\bibitem{Simonyan2014}
K.~Simonyan and A.~Zisserman.
\newblock Very deep convolutional networks for large-scale image recognition.
\newblock {\em CoRR}, abs/1409.1556, 2014.

\bibitem{Szegedy2016}
C.~Szegedy, S.~Ioffe, and V.~Vanhoucke.
\newblock Inception-v4, inception-resnet and the impact of residual connections
  on learning.
\newblock {\em CoRR}, abs/1602.07261, 2016.

\bibitem{Wang2017}
P.~Wang, P.~Chen, Y.~Yuan, D.~Liu, Z.~Huang, X.~Hou, and G.~W. Cottrell.
\newblock Understanding convolution for semantic segmentation.
\newblock {\em CoRR}, abs/1702.08502, 2017.

\bibitem{Wu2016}
Z.~Wu, C.~Shen, and A.~van~den Hengel.
\newblock High-performance semantic segmentation using very deep fully
  convolutional networks.
\newblock {\em CoRR}, abs/1604.04339, 2016.

\bibitem{Wu2016c}
Z.~Wu, C.~Shen, and A.~van~den Hengel.
\newblock Wider or deeper: Revisiting the resnet model for visual recognition.
\newblock {\em CoRR}, abs/1611.10080, 2016.

\bibitem{Xie2016}
S.~Xie, R.~Girshick, P.~Doll{\'{a}}r, Z.~Tu, and K.~He.
\newblock Aggregated residual transformations for deep neural networks.
\newblock {\em CoRR}, abs/1611.05431, 2016.

\bibitem{Xu+15}
B.~Xu, N.~Wang, T.~Chen, and M.~Li.
\newblock Empirical evaluation of rectified activations in convolutional
  network.
\newblock {\em CoRR}, abs/1505.00853, 2015.

\bibitem{Zagoruyko2016WRN}
S.~Zagoruyko and N.~Komodakis.
\newblock Wide residual networks.
\newblock In {\em (BMVC)}, 2016.

\bibitem{LSUNSeg17}
Y.~Zhang, H.~Zhao, and J.~Shi.
\newblock {LSUN}2017 segmentation challenge winning team {PSPN}et, July 2017.

\bibitem{zhao2016pspnet}
H.~Zhao, J.~Shi, X.~Qi, X.~Wang, and J.~Jia.
\newblock Pyramid scene parsing network.
\newblock {\em CoRR}, abs/1612.01105, 2016.

\bibitem{Zho+16}
B.~Zhou, H.~Zhao, X.~Puig, S.~Fidler, A.~Barriuso, and A.~Torralba.
\newblock Semantic understanding of scenes through the {ADE20K} dataset.
\newblock {\em CoRR}, abs/1608.05442, 2016.

\end{thebibliography}
}

\end{document}